\documentclass{article}
\usepackage{amsmath,graphicx,mlspconf}
\usepackage{xcolor}
\usepackage{xurl} 
\usepackage{hyperref} 

\usepackage{amsmath}

\usepackage{amsfonts}

\usepackage{amssymb}

\usepackage{stmaryrd}

\usepackage[mathscr]{eucal}

\usepackage{amsbsy}

\usepackage{bm}

\usepackage{bbding}

\usepackage{array}

\usepackage{tabularx}

\usepackage{ctable}

\usepackage[neveradjust]{paralist}

\usepackage{fancyvrb}

\usepackage{caption}
\usepackage{subfig}

\usepackage{placeins}

\usepackage{ifpdf}
\ifpdf
\DeclareGraphicsExtensions{.pdf,.png}
\else
\DeclareGraphicsExtensions{.eps}
\fi

\usepackage{algpseudocode}

\usepackage{xurl}

\newcommand{\email}[1]{\href{mailto:#1}{\nolinkurl{#1}}}

\newcommand{\Sec}[1]{\hyperref[sec:#1]{\S\ref*{sec:#1}}} 
\newcommand{\Section}[1]{\hyperref[sec:#1]{Section~\ref*{sec:#1}}} 
\newcommand{\AppFull}[1]{\hyperref[sec:#1]{Appendix~\ref*{sec:#1}}} 
\newcommand{\Eqn}[1]{\hyperref[eq:#1]{(\ref*{eq:#1})}} 
\newcommand{\Fig}[1]{\hyperref[fig:#1]{Figure~\ref*{fig:#1}}} 
\newcommand{\Tab}[1]{\hyperref[tab:#1]{Table~\ref*{tab:#1}}} 
\newcommand{\Thm}[1]{\hyperref[thm:#1]{Theorem~\ref*{thm:#1}}} 
\newcommand{\Cor}[1]{\hyperref[cor:#1]{Corollary~\ref*{cor:#1}}} 
\newcommand{\Alg}[1]{\hyperref[alg:#1]{Algorithm~\ref*{alg:#1}}} 
\newcommand{\Def}[1]{\hyperref[def:#1]{Definition~\ref*{def:#1}}} 

\newcommand{\Real}{{\mathbb R}}

\newcommand{\Tra}{^{{\sf T}}} 
\newcommand{\V}[1]{{\bm{\mathbf{\MakeLowercase{#1}}}}} 

\newcommand{\Oprod}{\circ} 

\newcommand{\M}[1]{{\bm{\mathbf{\MakeUppercase{#1}}}}} 
\newcommand{\MC}[2]{\V{#1}_{#2}} 

\newcommand{\T}[1]{\boldsymbol{\mathscr{\MakeUppercase{#1}}}} 

\newcommand{\norm}[1]{\left\lVert \, #1 \, \right\rVert}

\title{Knowledge-Guided Pattern Discovery via Coupled Tensor Factorizations}
\name{%
   Gaute Johannessen$^{\star}$%
   \qquad Geert Roelof van der Ploeg$^{\dagger}$%
   \qquad Evrim Acar$^{\dagger}$%
}
\address{%
   $^{\star}$ Faculty of Mathematics and Natural Sciences, University of Oslo, Oslo \\
   $^{\dagger}$ Department of Data Science and Knowledge Discovery, \\ Simula Metropolitan Center for Digital Engineering, Oslo %
}

\begin{document}

\maketitle

\begin{abstract}

\noindent In order to understand complex systems such as the human metabolome or human brain, different sensing technologies are used, generating complex data. These datasets are often multiway, i.e., with more than two axes of variation such as a \emph{subjects} by \emph{metabolites} by \emph{time} array. While tensor factorizations have successfully revealed interpretable patterns from such complex data, they have so far been mainly data-driven. On the other hand, there is more to data -- there are computational models (of these systems), which are rich sources of prior information. In this paper, we introduce a knowledge-guided approach that brings together data and computational models by jointly analyzing real data and simulated data (generated using a computational model) using coupled tensor factorizations with linear coupling. Our experiments on real metabolomics measurements  demonstrate that guiding the analysis of such noisy data with simulated data improves the pattern discovery performance while also revealing potential discrepancies between data and computational models.


\end{abstract}

\newcommand{\cem}[1]{\textcolor{blue}{cem: #1}}
\section{Introduction}
\label{sec:intro}
Complex systems such as the human metabolome (i.e., the complete set of small biochemical compounds in the body) or the human brain produce high-dimensional, noisy measurements that evolve over time and with significant variation between individuals. Such datasets can often be arranged as higher-order tensors, e.g., \textit{subjects} by \textit{metabolites} by \textit{time} tensors representing longitudinal metabolomics data \cite{Yan2024}, \textit{subjects} by \textit{voxels} by \textit{time} tensors in neuroscience \cite{MoAcAd25}. Tensor factorizations \cite{BaKo25} have been successfully used to reveal interpretable patterns from such data in  neuroscience \cite{MoAcAd25, DrZh25}, metabolomics \cite{Yan2024}, and microbiome data analysis \cite{Martino2021}.

While tensor factorizations have proven effective, existing approaches are predominantly data-driven and do not exploit the rich prior information encoded in computational models of these systems. Computational models, grounded in known biochemistry and physiology, encode dynamics that may be difficult to infer from noisy observations. In recent years, there have been significant efforts to develop knowledge-guided machine learning methods \cite{KaKe21,MoLi21,KaJi24,RuMa23} to incorporate prior information to improve scalability, generalizability, explainability, and with much smaller data sizes. While these methods hold great potential, their progress has been primarily limited to inference/prediction tasks and data-rich domains. Recent work has introduced goal-oriented tensor factorizations preserving underlying physical quantities of interest via penalty terms \cite{DuPh25}. Nevertheless, incorporation of such prior information from computational models in unsupervised settings where the goal is to reveal the underlying patterns (with high accuracy and reproducibility) and extract novel insights is relatively underdeveloped.

\begin{figure}[t]
  \centering
  \includegraphics[width=\linewidth, trim=10 10 10 10]{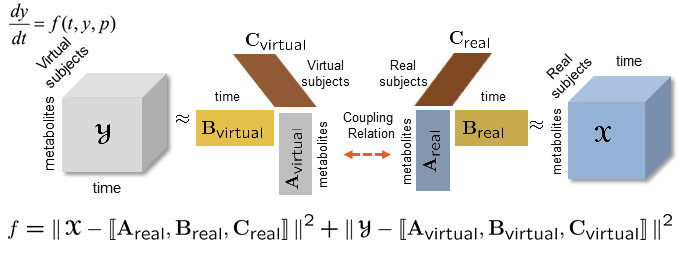}
  \caption{Coupled CP model jointly analyzing real data $\T{X}$ with simulated data $\T{Y}$, where the factor matrices in the metabolites mode, $\M {A}_{\text{virtual}}$ and $\M {A}_{\text{real}}$, are linked via a coupling relation.}
  \label{fig:coupled}
\end{figure}

In this paper, we introduce a novel knowledge-guided approach for pattern discovery bringing together data and computational models. Suppose we have real (observational) data collected from a dynamic system -- represented as a higher-order tensor, e.g.,
$\T{X}$ with modes: metabolites, time, subjects. Using a computational model of the system, we also generate simulated data, e.g., tensor $\T{Y}$ with modes: metabolites, time, virtual subjects.
We then jointly analyze $\T{X}$ and $\T{Y}$ coupled in the features (i.e., metabolites) mode using coupled tensor factorizations to guide the analysis of noisy real data with relatively clean simulated data (Fig. \ref{fig:coupled}). Coupled tensor factorizations offer a principled way to jointly analyze real and simulated data using tensor models, e.g., the CANDECOMP/PARAFAC (CP) model, while extracting latent factors in the coupled mode linked to each other via a coupling relation. Since sets of features in real and simulated data may only partially overlap, we make use of coupled tensor factorizations with linear coupling \cite{Schenker2021, ScWa25}. In this coupled setting, we consider different models for real data, i.e., CP and a more flexible PARAFAC2 model that accounts for individual variation by extracting subject-specific time profiles. 
Previously, real and simulated datasets were jointly analyzed \cite{Li2024}; however, unlike our approach, previous work was limited to a coupled CP model where only the matching features (e.g., metabolites) were included: only 6 out of the over 150 available in real data.
In this paper, we guide real data analysis with simulated data in a more realistic setting by including all features in the real data. Our experiments using coupled models relying on both CP and PARAFAC2-based approaches demonstrate that coupling real data analysis with simulated data consistently improves pattern discovery. 

\section{Methods}
\label{sec:dataandmethods}

We first discuss the tensor models used to extract patterns from the real data alone, and then introduce the coupled tensor factorization models used to jointly analyze real and simulated datasets.
\vspace{-0.1in}
\subsection{Tensor Factorizations}
The CP model \cite{Harshman1970, CarrollChang1970} approximates a higher-order tensor using a sum of rank-one tensors. An $R$-component CP model of a third-order tensor $\T{X} \in \mathbb{R}^{I \times J \times K}$ can be represented as
\begin{equation}
\label{eq:cp_def}
\small
    \T X \approx \sum_{r=1}^R {\V a_r \Oprod \V b_r \Oprod \V c_r} = \llbracket \M A, \M B, \M C \rrbracket,
\end{equation}
where $\Oprod$ is the outer vector product; $\M A \in \mathbb{R}^{I\times R}, \M B \in \mathbb{R}^{J\times R}, \M C \in \mathbb{R}^{K\times R}$ are the factor matrices in each mode. Under mild conditions, the CP model is unique up to permutation and scaling ambiguities \cite{BaKo25}, which facilitates interpretation. If $\T{X}$ is a \emph{metabolites} by \emph{time} by \emph{subjects} tensor, the $r$th component may reveal groups of metabolites in $\MC{A}{r}$, their temporal profile in $\MC{B}{r}$ and the corresponding subject stratification in $\MC{C}{r}$. The CP model can be fit to $\T{X}$ by solving the optimization problem:
\begin{equation}
\label{eq:cp_opt}
    \min_{\M A, \M B, \M C} \norm{\T X - \llbracket \M A, \M B, \M C \rrbracket}^2,
\end{equation}
where $\norm{\cdot}$ denotes the Frobenius norm.  

The PARAFAC2 model \cite{Harshman1972} relaxes the CP model by allowing the captured patterns in one mode to change across slices. An $R$-component PARAFAC2 model approximates $\T{X}$ as follows:
\begin{equation}
\label{eq:par2_def}
    \M X_k \approx \M A \M D_k \M B_k \Tra, \quad \M B_k\Tra \M B_k = \M \Phi, \quad \text{for } k = 1,\dots,K,
\end{equation}
where $\M X_k \in \mathbb{R}^{I \times J}$ corresponds to the $k$th frontal slice of $\T X$, $\M{D}_k \in \mathbb{R}^{R\times R}$ is a diagonal matrix with $\M{C}(k,:) = \text{diag}(\M{D}_k)$, i.e., the $k$th row of \M{C} has the diagonal entries of $\M{D}_k$. The cross product matrix $\M \Phi \in \mathbb{R}^{R\times R}$ is constant for all $k$, $k=1, ..., K$. Similar to CP, PARAFAC2 has uniqueness properties which facilitate the interpretation of factors \cite{kiersPARAFAC2PartDirectFitting1999}. In PARAFAC2,  \M{A} and \M{C} can reveal groups of metabolites and subject stratifications, as in the CP model. On the other hand, $\M B_k \in \Real^{J \times R}$ represents the time profiles for subject $k$. Instead of a shared set of time profiles, i.e., $\M B $ matrix in the CP model, PARAFAC2 allows each subject to have different time profiles -- which has the promise to reveal subject-specific differences \cite{Chatzis2026, ErCh26, MaCh17}.
When fitting a PARAFAC2 model, we solve the following optimization problem:
\begin{align}
\begin{split}
    \min_{\M A, \{\M B_k, \M D_k\}_{k \leq K} } &\sum_{k=1}^K{\norm{\smash{\M X_k - \M A \M D_k \M B_k\Tra}}^2} \\
       \text{s.t. \hspace{0.2in}} & \M B_k\Tra \M B_k = \M \Phi, \ \text{for } k = 1,\dots,K.
    \label{eq:parafac2}
\end{split}
\end{align}
\vspace{-0.2in}
\subsection{Coupled Tensor Factorizations}
To utilize the prior knowledge encoded in simulated data, real data tensor $\T{X}$ and simulated data tensor $\T{Y}\in \Real^{I_v \times J_v \times K_v}$  can be jointly analyzed using coupled tensor factorizations as in Fig. \ref{fig:coupled}. Coupled tensor factorizations are extensions of tensor factorizations to multiple datasets \cite{Acar2011, Acar2015, SoBaLa15}. 
For instance, $\T{X}$ and $\T{Y}$ coupled in the first mode, e.g., metabolites mode, can be jointly analyzed using a coupled CP model as follows:
\begin{align}
    \min_{\substack{\M A, \M B, \M C, \\ \M \Delta, \M A_v, \M B_v, \M C_v}}
    & \norm{\T X - \llbracket \M A, \M B, \M C \rrbracket}^2 \hspace{-0.02in}+ \hspace{-0.02in}\norm{\T Y - \llbracket \M A_v, \M B_v, \M C_v \rrbracket}^2 \nonumber\\
    \text{s.t. \hspace{0.2in} } & \M A = {\M H}_r\M \Delta \text{ and } \M A_v = {\M H}_v \M \Delta,
    \label{eq:cpcp}
\end{align}
where $\M{A}_v \in \Real^{I_v \times R}$, $\M{B}_v \in \Real^{J_v \times R}$ and $\M{C}_v \in \Real^{K_v \times R}$ are the factor matrices of the simulated data in each mode; ${\M H}_\text{r}$ and ${\M H}_\text{v}$ are transformation matrices defining the coupling relation, i.e., indicating the matching metabolites in each dataset, and $\M \Delta \in \Real^{I \times R}$ is the consensus matrix. Here we assume an $R$-component model for both datasets. When simulated data contains only a subset of the metabolites in the real data, we can define $\M{H}_r = \M{I} \in \mathbb{R}^{I\times I}$ and $\M{H}_v \in \mathbb{R}^{I_v\times I}$ with a single 1 in each row matching the shared metabolites in $\T{X}$ and all other entries zero. The coupling relation in (\ref{eq:cpcp}) is relevant for our application of interest.  More general linear coupling relations are also possible \cite{Schenker2021, ScWa25} but not considered here.

Similarly, $\T{X}$ and $\T{Y}$ can be jointly analyzed using a coupled PARAFAC2-CP model \cite{ScWa25} by solving the following optimization problem:
\begin{align}
\min_{_{\substack{\M A_v, \M B_v, \M C_v \\ \M \Delta,\M A, \{\M B_k, \M D_k\}_{k\leq K} }}}
       & \hspace{-0.2in}\sum_{k=1}^K\norm{\smash{\T X_k - \M A \M D_k \M B_k\Tra}}^2  \hspace{-0.06in}+\hspace{-0.05in} \norm{\T Y - \llbracket \M A_v, \M B_v, \M C_v \rrbracket}^2 \label{eq:parafac2cp} \\
    \text{s.t. \hspace{0.1in} } & \M A = {\M H}_r\M \Delta ,\ \M A_v = {\M H}_v \M \Delta, \nonumber\\
    &  \M B_k\Tra \M B_k = \M \Phi, \ \text{for } k = 1,\dots,K, \nonumber
\end{align}
where real data $\T{X}$ is approximated using a PARAFAC2 model to reveal subject-specific time profiles, $\T{Y}$ is approximated using a CP model, and $\M{H}_r$ and $\M{H}_v$ can be defined as in (\ref{eq:cpcp}).

\section{Experiments}

\subsection{Data}
In our experiments, we use two metabolomics datasets: a real dataset from a meal challenge study and a simulated dataset generated using a human metabolic model. Both are
arranged as \emph{metabolites} by \emph{time points} by \emph{subjects} tensors. The goal is to reveal underlying patterns characterizing metabolic responses of subjects to a meal and study whether those patterns (in time and metabolites mode) can stratify subjects according to body composition and insulin resistance measures \cite{Yan2024, Chatzis2026}.\\
\noindent \textbf{Real data} consists of measurements of samples collected during a meal challenge test from the COPSAC\textsubscript{2000} cohort \cite{Bisgaard2004} in which participants consumed a standardized meal after overnight fasting. Blood samples were taken at the fasting state (0hr) and seven time points after the meal (0.25hr, 0.5hr, 1hr, 1.5hr, 2hr, 2.5hr and 4hr), and measured using Nuclear Magnetic Resonance spectroscopy. In addition, we have measurements of insulin and C-peptide. Due to previously discovered sex-related differences in the metabolic response \cite{Yan2024}, we restricted our analysis to male subjects only. Four subjects were removed as outliers, as in previous studies \cite{Yan2024, Li2024, Chatzis2026}. We also removed seven subjects with missing insulin and C-peptide values across all time points. The data is a tensor of 161 metabolites $\times$ 8 time points $\times$ 133 subjects. 

\noindent \textbf{Simulated data} is from Li et al. \cite{Li2024} and was generated using a human whole-body metabolic model \cite{Kurata2021}, which is a kinetic model defined by ordinary differential equations with metabolites as variables and kinetic constants as parameters, capturing multi-organ metabolic dynamics after a meal. Fifty virtual subjects were generated by randomly perturbing kinetic parameters to introduce individual variation, yielding cleaner metabolic dynamics than noisy real measurements. The simulated data contains concentrations of six blood metabolites/hormones: insulin (Ins), glucose (Glc), pyruvate (Pyr), lactate (Lac), alanine (Ala) and $\beta$-hydroxybutyrate (Bhb), all of which are also available in the real data. These were measured at the same time points as in the real data, yielding a tensor of 6 metabolites $\times$ 8 time points $\times$ 50 virtual subjects.
\vspace{-0.1in}
\subsection{Experimental setup}
To assess the added value of joint analysis of real and simulated data, we compare the performance of (various) coupled models of real and simulated datasets with individual analysis of real data. We focus on the following settings:\\
\textbf{CP (T0-corrected) vs. Linearly Coupled CP - CP (T0-corrected)}: We compare the CP model of the real data with the coupled CP model of real and simulated data with linear coupling in the metabolites mode as in (\ref{eq:cpcp}). 
Datasets are T0-corrected, i.e., measurements at the first time point are subtracted from other time slices. Both datasets are centered across the subjects mode and scaled within the metabolites mode. We consider this setting since T0-correction was used in previous studies \cite{Yan2024, Li2024}. 

\noindent \textbf{CP vs. Linearly Coupled CP - CP}: We compare the CP model of the real data with the coupled CP model of real and simulated data with linear coupling in the metabolites mode as in (\ref{eq:cpcp}). When there is no T0-correction, datasets are nonnegative. We add nonnegativity constraints to all modes in (\ref{eq:cpcp}), and scale the datasets within the metabolites mode. 

\noindent \textbf{PARAFAC2 vs. Linearly Coupled PARAFAC2 - CP}: 
Finally, we compare a PARAFAC2 model of the real data with a PARAFAC2 model of the real data coupled with a CP model of the simulated data as in (\ref{eq:parafac2cp}). We analyze the real data using a PARAFAC2 model to account for individual differences in time in terms of metabolic response to the meal.  Again, there is no T0-correction and datasets are nonnegative. Nonnegativity constraints are added to all modes in (\ref{eq:parafac2cp}), and datasets are scaled within the metabolites mode.

For all models, missing entries in real data (about 0.01\%) are imputed by the mean of the corresponding subject and time fibers. For linearly coupled CP-CP models, we have a ridge penalty on all modes (which does not improve performance for other models; see Supplementary Material [SM]).

We assess the performance of the models in terms of pattern discovery based on the correlation between factors extracted from the subjects mode and additional subject information, i.e., body composition and insulin resistance measures including homeostatic model assessment for insulin resistance (HOMA-IR), muscle to fat ratio, fat percentage, muscle mass, weight, Body Mass Index (BMI), waist circumference, waist to height ratio, fat mass, fat mass index, and fat-free mass index (FFMI). Detailed descriptions are in \cite{Yan2024}.

\noindent{\bf{Model selection.}} We select the number of components ($R$) when fitting (coupled) tensor factorizations by considering reproducibility and replicability of the components as well as their biological interpretation.
Reproducibility measures the ability to recover a unique solution, while replicability measures whether similar patterns are found across overlapping subsets of the data \cite{MoAcAd25}. 
Procedural details, reproducibility and replicability plots are included in the SM.

\noindent \textbf{Implementation details.}
The CP models are fitted using \texttt{cp\_wopt} (from the Tensor Toolbox - version 3.7) using the limited memory BFGS algorithm. We use the Alternating Optimization (AO) - Alternating Direction Method of Multipliers (ADMM) framework for PARAFAC2 \cite{Roald2022} and for linearly coupled tensor models \cite{Schenker2021,ScWa25}. For all models, we use 100 random initializations and select the run with the lowest loss. The convergence is assessed using an absolute tolerance of $10^{-8}$ and a relative tolerance of $10^{-6}$ on the loss. The code and SM are available at: \url{https://github.com/GauteJ1/CoupledModelsProject/}.

\section{Results}
\label{sec:results}

\begin{figure}[t]
  \centering
  \includegraphics[width=\linewidth, trim=0 0 0 0]{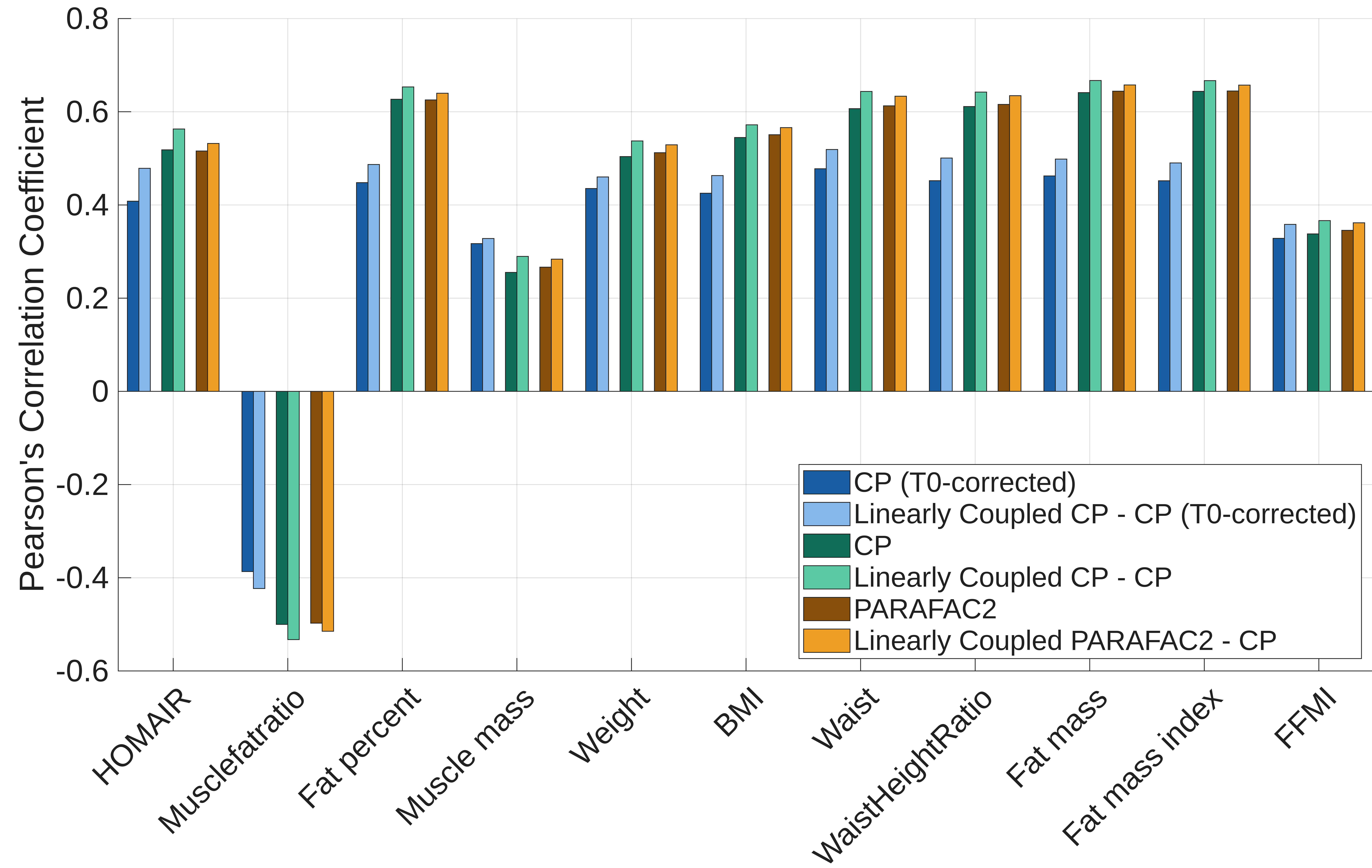}
  \caption{Correlations between subject coefficients and body composition and insulin resistance measures, for the component with the highest correlations in each model. Models are colored in pairs, comparing uncoupled (dark) and linearly coupled (light) factorizations.}
  \label{fig:correlations}
\end{figure}

Our results demonstrate that jointly analyzing noisy real data with simulated data improves the recovery of a biologically meaningful component. The clean simulated data has an underlying component modeling co-varying insulin and glucose levels. 
When coupled with real data, the simulated data guides the analysis and enables recovery of this component known to be strongly linked to a BMI-related phenotype \cite{Li2024}.
In uncoupled models of real data, this component is either not captured because other metabolites dominate the model or is noisier. Therefore, the patterns extracted by the coupled models have consistently stronger correlations with body composition and insulin resistance measures (Fig. \ref{fig:correlations}). 

\noindent \textbf{CP (T0-corrected) vs. Linearly Coupled CP-CP (T0-corrected)}:
A 2-component CP model of T0-corrected real data reveals two metabolite factors, with the second one capturing a BMI-associated pattern \cite{Yan2024}. However, this pattern is mainly dominated by metabolites other than insulin, glucose (see Supp Fig. 1, comp 2). 
When the real data is analyzed together with simulated data using a 3-component linearly coupled CP-CP model, we observe that coupling forces a pattern in the simulated data mainly capturing insulin and glucose into the model of the real data (see Supp Fig. 4, comp 3), yielding stronger BMI-associated correlations.

\noindent \textbf{CP vs. Linearly Coupled CP-CP}:
A 6-component CP model of the real data with nonnegativity constraints reveals several factors relevant for BMI-related variables. Among them is a clear insulin-dominated factor, but glucose is not modeled well and has a small coefficient (Supp Fig. 7, comp 5). On the other hand, when real data is jointly analyzed with simulated data using a 6-component linearly coupled CP-CP model, the model finetunes this component, increasing the glucose coefficient while suppressing the other metabolites, resulting in a purer insulin/glucose component with stronger correlations with the variables of interest (Supp Fig. 10, comp 5).

\begin{figure*}[!t]
  \centering
  \includegraphics[width=0.76\textwidth]{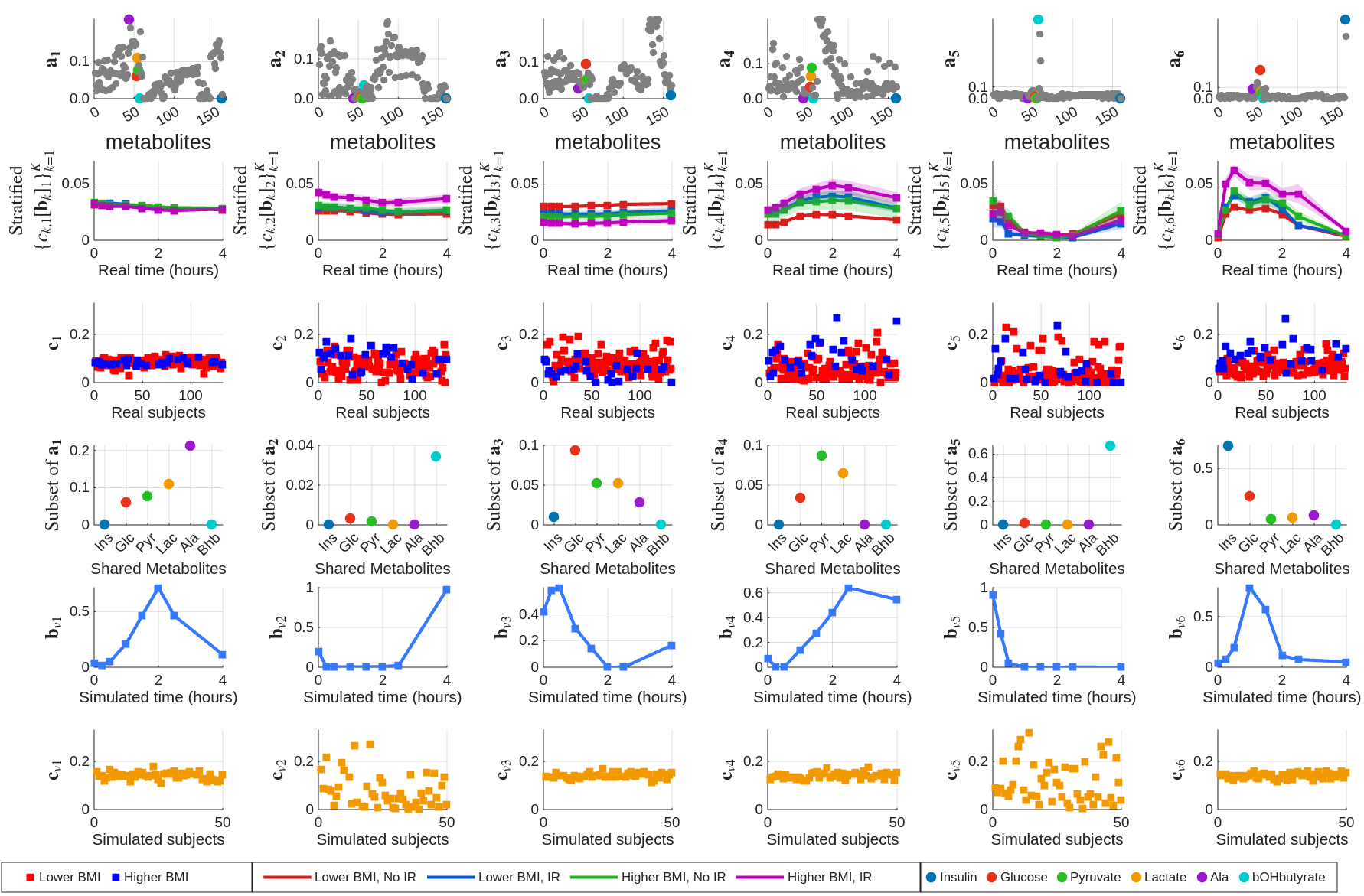}
  \caption{6-component linearly coupled PARAFAC2-CP model of real and simulated data, showing the linearly coupled metabolite factors ($\mathbf{a}$), subject factors ($\mathbf{c}$), simulated time profiles ($\mathbf{b}_v$) and virtual subject factors ($\mathbf{c}_v$). Time profiles of the real data ($\mathbf{b}_k$) scaled by subject coefficients are shown using the mean (and standard error of the mean) of stratifications based on BMI and Insulin Resistance (IR), i.e., lower BMI: BMI $<25$, higher BMI: BMI $\geq 25$; NoIR: HOMA-IR$\leq 2.91$, IR: HOMA-IR $> 2.91$.}
  \label{fig:PARAFAC_CP_model}
\end{figure*}

\noindent \textbf{PARAFAC2 vs. Linearly Coupled PARAFAC2-CP}:
The 6-component PARAFAC2 model of real data similarly recovers an insulin-dominated factor with a small glucose coefficient (Supp Fig. 13, comp 6). When the real data is jointly analyzed with simulated data using the 6-component linearly coupled PARAFAC2-CP model, again a cleaner insulin/glucose pattern (comp 6 in Fig. \ref{fig:PARAFAC_CP_model}) is captured yielding stronger correlations with BMI-related variables. Fig. \ref{fig:PARAFAC_CP_model} shows the factors of the linearly coupled PARAFAC2-CP model, which allows for subject-specific time profiles in the real data. We only include the factors of this model here. All other models are in SM.

\begin{figure*}[!t]
  \centering
  \includegraphics[width=0.76\textwidth]{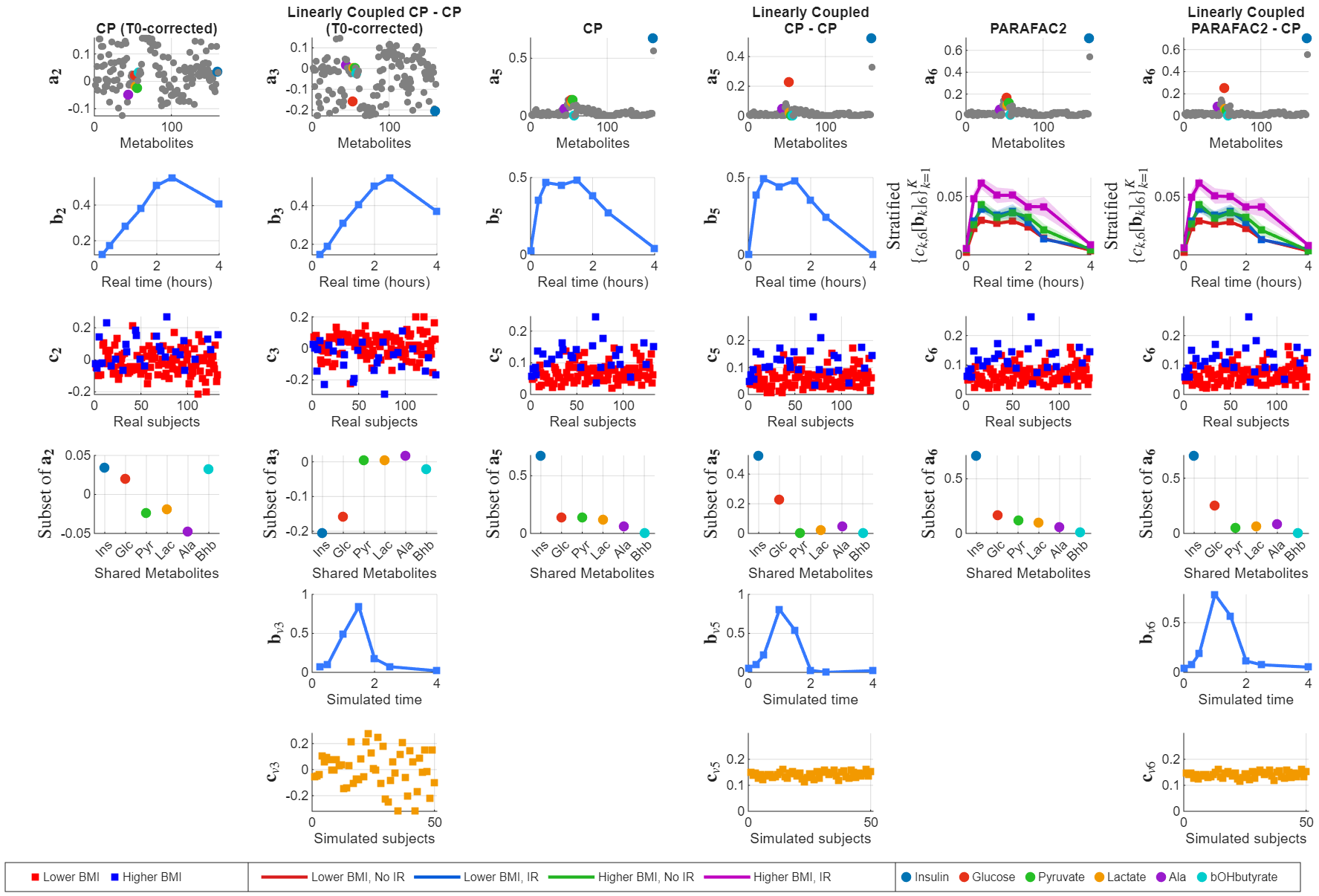}
  \caption{The component of interest (i.e., the one modeling insulin/glucose) for each model, showing the linearly coupled metabolite factor ($\mathbf{a}$), time profile ($\mathbf{b}$), subject factor ($\mathbf{c}$), simulated time profile ($\mathbf{b}_v$) and virtual subject factor ($\mathbf{c}_v$). For PARAFAC2 and coupled PARAFAC2-CP, subject-specific time profiles scaled by the subject coefficients are shown using the mean (and standard error of the mean) of stratifications based on BMI and Insulin Resistance (IR) groups.}
  \label{fig:ins_glc_comparison}
\end{figure*}

Fig. \ref{fig:ins_glc_comparison} highlights the insulin/glucose component across all models, illustrating how coupling consistently produces a cleaner pattern in terms of modeling insulin/glucose relation. 
Note that the simulated time profile of this component shows a sharper peak around 1 hour post-meal than observed in the real data, where the time pattern is much broader using CP and shows much more individual variation using PARAFAC2.
This demonstrates a potential discrepancy between the real data and the computational model. Furthermore, coupled models (via the factors extracted from the subjects mode) show that there is much less individual variation in the simulated data compared to real data.

\vspace{-0.1in}
\section{Conclusion}
\label{sec:conclusion}
In this paper, we have introduced a knowledge guided approach for interpretable pattern discovery from complex data. We demonstrate that coupled tensor factorizations with linear coupling constraints are effective tools for bringing together data and computational models via joint analysis of real and simulated data.
Our experiments on metabolomics data demonstrate that the proposed approach consistently provides cleaner patterns compared to the analysis of only real data. Besides, our results show that the coupled framework can reveal potential mismatches between real data and computational models. Future work should focus on whether conflicting information in simulated and real data can also be reliably extracted via coupled tensor factorizations, for instance, through unshared factors. We also plan to apply the proposed approach in neuroscience, where computational models and noisy high-dimensional measurements are common.

\vspace{-0.1in}
\section{Acknowledgements}
We thank the children and families of the COPSAC\textsubscript{2000} cohort for their contribution, and the clinical team at COPSAC for conducting the clinical study.

\bibliographystyle{IEEEbib}
\bibliography{strings,refs}

\end{document}